\documentclass{article} 
\usepackage{iclr2027_conference,times}

\usepackage{amsmath,amsfonts,bm}

\def\eqref#1{equation~\ref{#1}}

\def\1{\bm{1}}

\DeclareMathAlphabet{\mathsfit}{\encodingdefault}{\sfdefault}{m}{sl}
\SetMathAlphabet{\mathsfit}{bold}{\encodingdefault}{\sfdefault}{bx}{n}

\usepackage{hyperref}
\usepackage{url}
\usepackage{graphicx}
\usepackage{wrapfig}   

\usepackage{booktabs}
\usepackage{multirow}
\usepackage{makecell}
\usepackage{pifont}
\usepackage{capt-of}

\usepackage{listings, xcolor}
\lstdefinestyle{promptstyle}{
  basicstyle=\scriptsize\ttfamily, breaklines=true, breakindent=0pt,
  columns=fullflexible, frame=single, framesep=4pt, rulecolor=\color{gray!50},
  backgroundcolor=\color{gray!5}, xleftmargin=2pt, xrightmargin=2pt,
  aboveskip=8pt, belowskip=8pt,
}

\newcommand{\xmark}{\ding{55}}
\newcommand{\cmark}{\ding{51}}

\title{Ego-Forge: Text and Geometric-Attention Free Exo-to-Egocentric Video Generation}

\author{Mohammad Mahdi\thanks{Corresponding author} \quad Luc Van Gool \quad Danda Pani Paudel  \\
INSAIT, Sofia University “St. Kliment Ohridski” \\
\texttt{\{firstname.lastname\}@insait.ai}
}

\iclrfinalcopy 
\begin{document}

\maketitle
\begin{figure}[h]
    \centering
    \includegraphics[width=\linewidth]{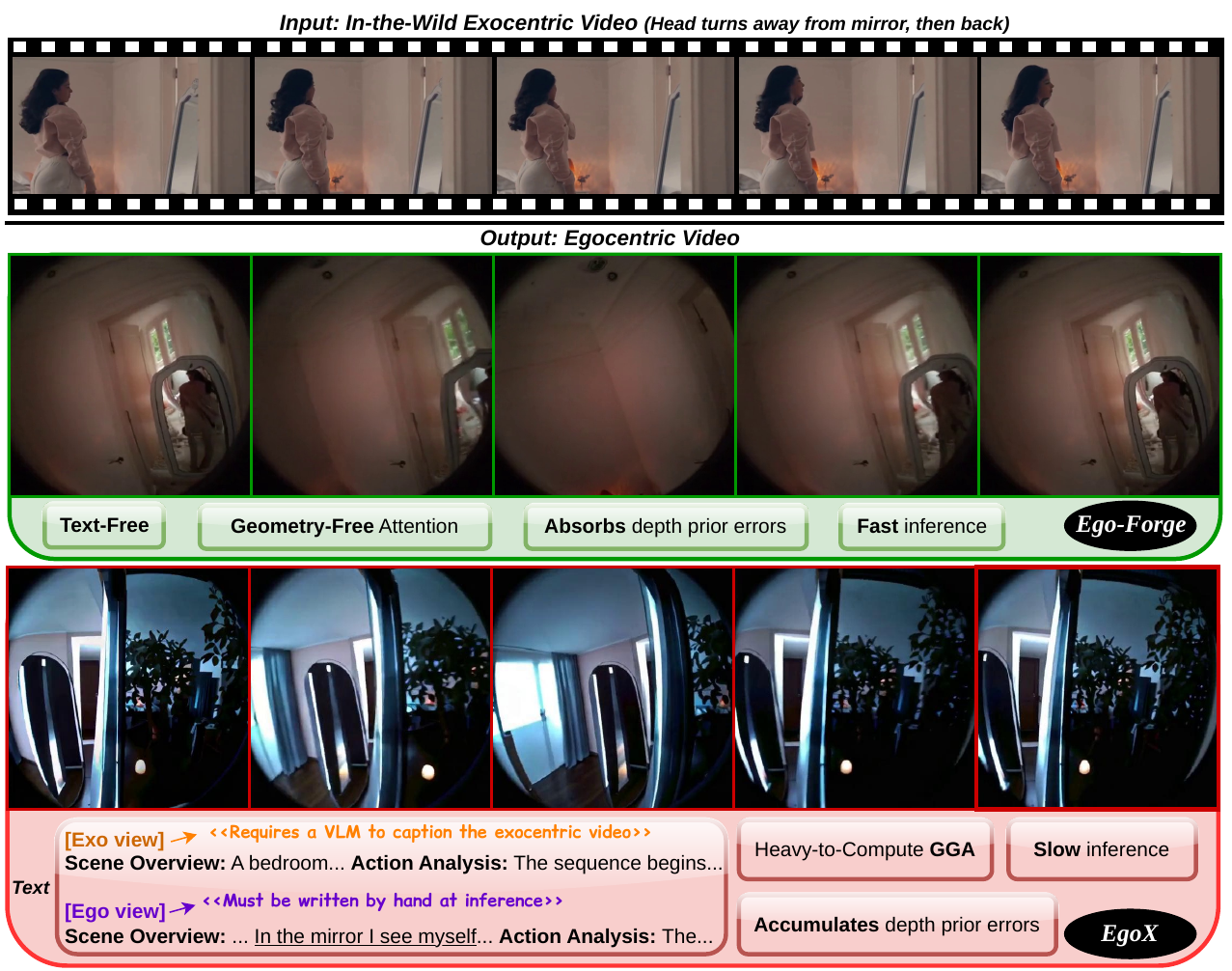}
    \caption{\textbf{Ego-Forge generates egocentric video without text conditioning or geometry-guided attention.} \textit{Top:} An in-the-wild exocentric clip in which the subject turns away from a mirror and then back. \textit{Middle:} Our method follows the head motion and renders the mirror and its reflection, although the exocentric camera never observes the reflected content and no caption is provided at inference. \textit{Bottom:} EgoX, conditioned on captions of both views, including an explicit description of the reflection, and Geometry-Guided Self-Attention (GGA), fails to recover the reflection.}

    \label{fig:teaser}
\end{figure}
\begin{abstract}
Exo-to-egocentric video generation aims to synthesize what a person sees from their own viewpoint given third-person footage and a target head trajectory. The task requires transferring appearance and semantics across large viewpoint changes while hallucinating content never observed by the exocentric camera. Existing approaches either impose additional input requirements, such as a ground-truth initial egocentric frame or multiple synchronized exocentric views, or remain limited to category-specific settings. EgoX~\citep{kang2026egox} is the first to address cross-activity and in-the-wild generalization, but requires a human-provided caption of the non-existent egocentric view at inference and introduces a computationally expensive geometry-guided attention bias that can propagate reconstruction errors and suppress textual and visual context (Figure~\ref{fig:teaser}). We therefore propose \textbf{Ego-Forge}, a caption-free and bias-free framework for exo-to-egocentric generation. It introduces \textit{Dynamic Captioning}, which derives conditioning tokens directly from the model's hidden states and adapts them to the diffusion timestep and network depth, replacing external text conditioning. By scaling training by an order of magnitude and using all available exocentric viewpoints, Ego-Forge learns cross-view correspondence implicitly and eliminates the need for geometry-guided attention, requiring only a lightweight depth prior. Ego-Forge achieves state-of-the-art performance on Ego-Exo4D~\citep{grauman2024ego}, runs $4.3\times$ faster end-to-end, requires no external annotation at inference, and generalizes to in-the-wild scenes, including cases where over-reliance on geometry blocks appearance inference. Our model and source code will be made publicly available.

\end{abstract}

\section{Introduction}

Exo-to-egocentric video generation aims to synthesise what a person sees from their own viewpoint, given third-person footage of them acting and a target camera trajectory for their head. Beyond video synthesis, the task probes visual understanding: the model must infer scene appearance and semantics from an exocentric view, transfer them across a large viewpoint change, and maintain consistency with the specified trajectory, while hallucinating substantial portions of the egocentric view that are never observed by the exocentric camera. This capability has applications in augmented and virtual reality~\citep{engel2023project}, embodied learning~\citep{kim2024openvla,brohan2023rt,nair2022r3m, ma2022vip}, and egocentric training data generation~\citep{tran2026egoexo}, where paired exocentric-egocentric recordings are expensive to collect.

Existing methods make the problem tractable by constraining the setting: EgoExo-Gen~\citep{xu2025egoexo} assumes the ground-truth first egocentric frame is available, and Exo2Ego-V~\citep{liu2024exocentric} and Exo2EgoSyn~\citep{mahdi2025exo2egosyn} require four synchronised exocentric cameras. Syn2Seq-Forcing~\citep{mahdi2026synchrony} avoids such requirements, reformulating the task as continuous sequence modelling by interpolating between the two views. All of them, however, remain category-specific, requiring separate finetuning per activity class. A recent method, EgoX~\citep{kang2026egox}, is the first to \emph{generalise across activities}. It adapts a pretrained video diffusion model using three conditioning signals: (i) an egocentric prior obtained through 3D reconstruction and reprojection along the target trajectory, (ii) a textual \emph{caption of the egocentric view}, and (iii) a geometry-guided attention (GGA) bias.

We argue that requiring a caption of the \emph{non-existent} egocentric view, together with the issues of the GGA bias discussed below, makes this approach impractical. First, the caption must be provided manually at inference, since the target egocentric view does not yet exist. Second, the GGA bias introduces three additional limitations. (i) It incurs high computational cost because it is computed over every query-key pair, making both training and inference expensive. (ii) It depends on the 3D reconstruction, causing reconstruction errors to propagate directly into the attention bias and the generated video. (iii) It suppresses visual and textual context, and is not compatible when the appearance change over the viewpoints (due to its over-reliance on the geometry alone).  Figure~\ref{fig:teaser} illustrates the latter case, where the reflection of the person in the mirror, which appears differently in different viewpoints, is not inferred, despite being described in the text (which gets suppressed by the GGA). As shown in the Appendix~\ref{app_teaser}, removing GGA recovers the reflection (yet without producing the correct ego view). We therefore propose \textit{\textbf{Ego-Forge}}, which eliminates both textual and geometric conditioning, avoiding their associated costs and failure modes.

To avoid the ego-view's caption, \textbf{\textit{Ego-Forge}} introduces \emph{Dynamic Captioning}: a lightweight module produces the conditioning tokens directly from the model's own hidden states, so the cross-attention receives a description derived from the exocentric video rather than from text. It is conditioned on the diffusion timestep and block index, allowing its output to adapt as the egocentric view emerges during denoising and the representation evolves across network depth. Note that no text is generated in this process. The name only reflects that the module serves the role a caption would otherwise play when conditioning the backbone foundation model. On the other hand, we let the model implicitly learn cross-view correspondence and visual context from data, rather than imposing them through a hand-computed geometric prior. Removing the expensive GGA enables efficient end-to-end training on an order of magnitude more data from all exocentric viewpoints~\footnote{In the Ego-Exo4D dataset, each egocentric video is paired with 4 exocentric views. While EgoX selects the best camera view during training, we leverage all views.
}.  Trained this way, the model learns to tolerate noisy reconstructions rather than inherit their errors. It therefore requires only the exocentric frames and a lightweight depth prior, eliminating both the captioning model at training time and human annotation at inference.

Our contributions are:

\begin{itemize}
    \item \textbf{Dynamic Captioning.} A lightweight module that replaces the text stream, producing conditioning tokens directly from the model's own hidden states and adapting them to the denoising step and the depth at which they are read. It removes the need for a human-written caption at inference.

    \item \textbf{A caption-free, bias-free pipeline.} We show that the geometry-guided attention bias can be dropped entirely when training is scaled to all exocentric viewpoints rather than only the best-reconstructed ones, and that a lightweight monocular depth model suffices in place of a heavy reconstruction. This removes the model's dependence on reconstruction quality.

    \item \textbf{State-of-the-art quality, $4.3\times$ faster.} Ego-Forge outperforms the state of the art on standard metrics on the Ego-Exo4D~\citep{grauman2024ego} benchmark while being $4.3\times$ faster end-to-end, and generalises to in-the-wild footage, including cases where over-reliance on geometry blocks appearance inference. 
\end{itemize}

\section{Related Work}

\paragraph{Diffusion-based video generation.}
Diffusion models trained at scale have rapidly improved the quality and temporal
coherence of video generation~\citep{wan2025wan,huang2024learning,liu2025control,liu2025diverse,pan2025earthsynth,zhang2025show,yang2024cogvideox}.
Beyond an initial text prompt, a line of work introduces explicit camera control,
either by conditioning on trajectories~\citep{he2024cameractrl, wang2024motionctrl, yang2024direct}
or by manipulating temporal attention within a pretrained
generator~\citep{bai2025recammaster}. A complementary direction maintains an explicit 3D
representation during generation to keep content consistent across
viewpoints~\citep{ren2025gen3c, li2025vmem, yu2025trajectorycrafter}. These methods synthesise video
along a continuous camera path, but do not transform the viewpoint of an existing
video: the cross-view setting, where the source and target cameras differ by a large
and discontinuous pose change, remains largely unexplored.

\paragraph{Ego--exo cross-view generation.}
Paired ego--exo datasets~\citep{grauman2024ego, huang2024egoexolearn,grauman2022ego4d} have enabled a body of work on
cross-view {understanding}, including object correspondence~\citep{fu2025objectrelator} and visual question
answering~\citep{he2025egoexobench, lee2025towards}. Generation is considerably harder, since the
model must synthesise the large portion of the egocentric view that the exocentric
camera never observes, and existing methods make the problem tractable by adding
constraints. PMYS~\citep{luo2024put} first predict hand trajectories and then generate the
egocentric video conditioned on them. EgoWorld~\citep{park2025egoworld} reprojects a point
cloud built from exocentric depth and 3D hand pose, then inpaints the result. Grounded-Exo2Ego~\citep{wang2026grounded} uses a dual-branch video diffusion model that combines geometric anchoring from 3D reconstruction with semantic grounding for challenging regions.
EgoExo-Gen~\citep{xu2025egoexo} conditions on an action description together with the
ground-truth first egocentric frame. Exo2Ego-V~\citep{liu2024exocentric} requires four
synchronised exocentric cameras and a PixelNeRF-based representation, while
Exo2EgoSyn~\citep{mahdi2025exo2egosyn} repurposes a large pretrained video generator but relies
on a single predicted egocentric frame to guide generation and applies camera control
per frame despite the backbone's temporally coupled attention. Syn2Seq-Forcing~\citep{mahdi2026synchrony}
takes a different view of the problem, identifying the synchronisation-induced
discontinuity between the two views as the central difficulty and interpolating between
source and target videos so that the pair forms a single continuous sequence, which a
diffusion sequence model~\citep{song2025history} can then follow. However, these methods remain category-specific, requiring separate finetuning per activity class.

\paragraph{EgoX and its dependencies.}
EgoX~\citep{kang2026egox} is the first method in this setting to generalise across activities.
It adapts a pretrained video diffusion model~\citep{wan2025} with LoRA~\citep{hu2022lora} and conditions it on three
signals: an egocentric prior obtained by reconstructing the scene with a monocular
depth model~\citep{huang2025vipe} and reprojecting it along the target trajectory, a Geometry-Guided
Self-Attention (GGA) bias that steers attention toward spatially corresponding
regions, and a textual caption describing both exo and egocentric views. Two of these limitations are the focus of our work. First, the caption must describe a view that does not yet exist. Second, the GGA bias is applied to every (query, key) pair, causing reconstruction errors to propagate directly into the attention computation rather than being washed out. We show that both requirements can be eliminated.

\paragraph{Replacing text conditioning in pretrained generators.} Adapting a text-conditioned generator to a setting where captions are unavailable requires producing tokens in the space expected by its cross-attention layers. Prior approaches introduce additional learned modules throughout the backbone to provide such conditioning~\citep{alayrac2022flamingo,ye2023ip}. In a video diffusion backbone with many blocks---40 in the case of Wan2.1~\citep{wan2025}---this can require a separate set of parameters at multiple depths. Our Dynamic Captioning instead uses a single resampler shared across the network, conditioned on the diffusion timestep and block index, recovering depth-specific behaviour at a fraction of the parameter cost (Section~\ref{sec:stage2}).

\section{Method}
\begin{figure}[t]
    \centering
    \includegraphics[width=\linewidth]{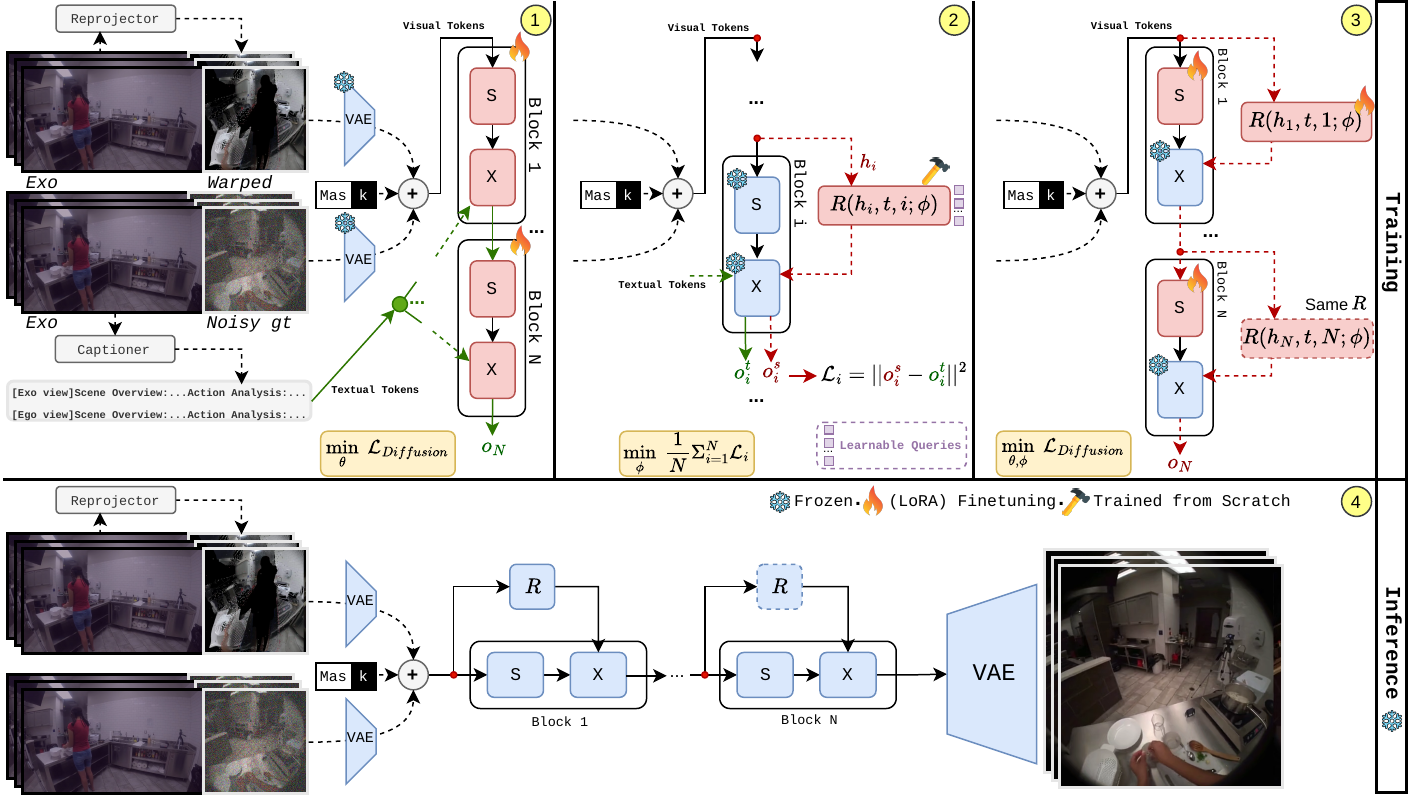}
    \caption{\textbf{Overview of Ego-Forge.} Throughout, $S$ denotes self-attention and $X$ cross-attention. \textbf{(1)~Text-conditioned adaptation.} We adapt the pretrained video diffusion backbone using all available exocentric viewpoints, conditioned on the exocentric video, reprojected egocentric prior, and a textual caption, without geometric attention bias. \textbf{(2)~Learning Dynamic Captioning.} A single resampler $R$, shared across all $N$ blocks, learns to replace the textual conditioning using the hidden state $h_i$, block index $i$, and diffusion timestep $t$. Here, the superscripts $^t$ and $^s$ denote the teacher and student, respectively. \textbf{(3)~Joint finetuning.} The caption is removed and the model is further finetuned with Dynamic Captioning and self-attention under the diffusion objective. \textbf{(4)~Inference.} Only the exocentric video and reprojected prior are required; $R$ provides the conditioning throughout the network without a caption.}

    \label{fig:framework}
\end{figure}

Ego-Forge, as illustrated in Figure~\ref{fig:framework}, is built in three training stages, followed by a caption-free inference.
Stage 1 (Section~\ref{sec:stage1}) adapts a pretrained video diffusion model~\cite{wan2025} to the
exo-to-ego task, conditioned on the exocentric frames, a
reprojected egocentric prior and a textual caption but without any geometric
attention bias, relying instead on large-scale training to let the model learn the
correspondence between the two views. Stage 2 (Section~\ref{sec:stage2}) freezes that model and trains a single Dynamic Captioner, $R(.)$, to reproduce what the frozen cross-attention would have produced from the caption, using the model's own hidden states as its input. Stage-3 (Section~\ref{sec:stage3}) removes the caption entirely and finetunes the Dynamic Captioner together with the self-attention layers under the diffusion objective alone. At inference (Section~\ref{sec:inference}) the model requires
only the exocentric video and a target camera trajectory: no caption is written, and no geometric attention bias is involved.

\subsection{Stage 1: Text-Conditioned Adaptation}
\label{sec:stage1}

\paragraph{Conditioning.}
We place the exocentric and egocentric views on a single canvas~\citep{kang2026egox}
so that a pretrained video generator can attend over both jointly. Given an exocentric
clip $V^{\text{exo}} \in \mathbb{R}^{F \times 3 \times H \times W_e}$ and a target head
trajectory $\{P_f\}_{f=1}^{F}$, we estimate scene geometry with a monocular depth
model~\citep{wang2025moge2} and reproject it along the trajectory to obtain an egocentric prior
$V^{\text{prior}} \in \mathbb{R}^{F \times 3 \times H \times W_g}$. The two are
concatenated along width and encoded with the frozen VAE, giving a latent canvas in
which the left $w_e$ columns carry the exocentric view and the right $w_g$ columns the
egocentric half to be generated.
Let $V^{\text{ego}} \in \mathbb{R}^{F \times 3 \times H \times W_g}$ denote the ground-truth egocentric video,  $x_0 = \mathcal{E}([\,V^{\text{exo}} \,\|\, V^{\text{ego}}\,])$ denote the latent canvas
of the ground-truth pair, and $x_t = (1-\sigma_t)\,x_0 + \sigma_t\,\epsilon$ be its noised
counterpart at timestep $t$ under the flow-matching schedule, with
$\epsilon \sim \mathcal{N}(0, I)$. The transformer input concatenates $x_t$, a binary
mask $m$ that is one over the exocentric columns and zero over the egocentric ones, and
the conditioning latent $c$, along the channel axis:
\begin{equation}
    z_t = \big[\, x_t \;\|\; m \;\|\; c \,\big], \qquad
    c = \mathcal{E}\big(\,[\,V^{\text{exo}} \,\|\, V^{\text{prior}}\,]\,\big).
\end{equation}

Text conditioning is supplied as in prior work: a caption describing both views is encoded with the frozen text encoder and read by the cross-attention layers. Unlike prior work,we generate these captions with Qwen3-VL-32B-Instruct~\citep{Qwen3-VL}, an open-weight VLM, rather than a paid API, which is what makes captioning the full training set feasible.

\paragraph{No geometric attention bias.}
Prior work adds a heavy-to-compute geometry-guided bias to the self-attention layers, steering each
egocentric query toward the exocentric regions its reconstruction places nearby. The bias is computed from the same reconstruction that produces the prior,
so wherever that reconstruction is wrong the bias is wrong in the same place, and
because it suppresses rather than reweights, the error is carried into the attention
rather than averaged away. We omit
this term. Removing it leaves the correspondence to be learned, which
requires more data than a geometric prior would: we therefore train on all available
exocentric viewpoints rather than restricting to the best-reconstructed camera per take,
yielding roughly $25\times$ more clips than prior work.

\paragraph{Training.}
Only the self-attention and cross-attention projections are adapted, with LoRA of rank
$r$, together with the input patch embedding, which must accommodate a conditioning
signal that now spans a stitched canvas rather than a single view. The backbone is
otherwise frozen. We train with the flow-matching objective, supervising only the
egocentric half of the canvas:
\begin{equation}
    \mathcal{L}_{\text{Diffusion}} =
    \big\| \,\epsilon_\theta(z_t, t, \tau)\big|_{\text{ego}} - v\big|_{\text{ego}} \,\big\|^2 ,
\end{equation}
where $\tau$ is the encoded caption and $v$ the flow-matching target. The exocentric
half is given content and is excluded from the loss.

\subsection{Stage 2: Learning the Dynamic Captioner}
\label{sec:stage2}

\paragraph{Rationale.}
The caption used during training is unavailable at inference because the target egocentric view does not yet exist. Removing the caption stream entirely, however, would discard a pathway the backbone was pretrained to use, leading to degraded performance (Table~\ref{tab:abl_conditioning}). We therefore retain the cross-attention pathway but replace its textual input with conditioning derived from the model's own hidden states. Our Dynamic Captioner $R(\cdot)$ reads the hidden state $h_i$ at block $i$ and is conditioned on both $i$ and the diffusion timestep $t$, allowing its output to adapt as the representation evolves across depth and the egocentric view emerges during denoising. Rather than using a separate module at each block, we share a single captioner across the network and provide $i$ and $t$ as inputs, achieving depth- and timestep-specific conditioning with a single set of parameters.

\paragraph{Architecture.}
\begin{wrapfigure}{R}{0.3\textwidth}
    \vspace{-12pt}
    \includegraphics[width=0.3\textwidth]{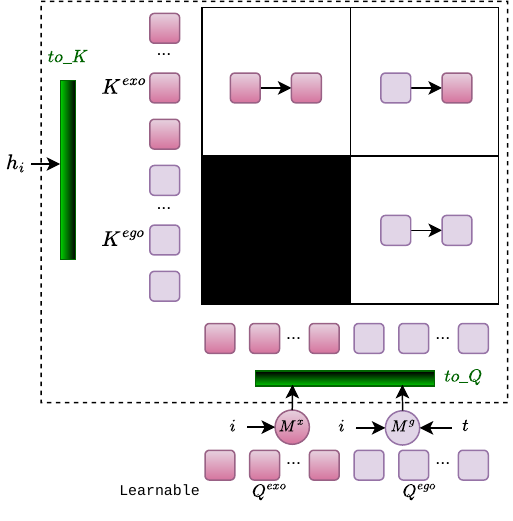}
    \vspace{-7pt}
    \caption{\textbf{R's attention structure.} $M^x$ and $M^g$ apply Eq~\ref{eq:4}.
The black block masks exo queries from ego keys.}
    \label{fig:resampler}
\end{wrapfigure}

$R$ is a cross-attention module with learned queries. It maintains two sets of queries,
$Q^{\text{exo}}$ and $Q^{\text{ego}}$ of $n$ tokens each, mirroring the two-part structure
of the captions used in Stage-1: $Q^{\text{exo}}$ attends only to the exocentric tokens of
$h_i$, while $Q^{\text{ego}}$ attends to the full canvas. Each set is modulated by a scale--shift transform before attending, but the two are
conditioned differently: $Q^{\text{ego}}$ on both the block index and the timestep, and
$Q^{\text{exo}}$ on the block index alone (Figure~\ref{fig:resampler}). The exocentric half is given content, so what
the model should extract from it depends on the depth at which it is read but not on how
noisy the canvas currently is. The outputs are concatenated and projected, giving $2n$
conditioning tokens --- matching the sequence length the cross-attention was pretrained
on:
\begin{equation}
    \hat{\tau}_i = W_o \big[\, \mathrm{Attn}(\tilde{Q}^{\text{exo}}, h_i^{\text{exo}}) \;\|\;
                               \mathrm{Attn}(\tilde{Q}^{\text{ego}}, h_i) \,\big],
\end{equation}
\begin{equation}
    \tilde{Q}^{\text{exo}} = Q^{\text{exo}} \odot (1 + \gamma_i) + \beta_i,
    \qquad
    \tilde{Q}^{\text{ego}} = Q^{\text{ego}} \odot (1 + \gamma_{i,t}) + \beta_{i,t} .
    \label{eq:4}
\end{equation}
$W_o$ is zero-initialised, so at the start of training, $R$ contributes nothing and the
model behaves exactly as it would with no conditioning.

\paragraph{Training.}
The Stage-1 weights are loaded and frozen, including the cross-attention layers. Freezing
them is what makes the objective well posed: if they could adapt, the loss could be reduced
by moving the cross-attention toward whatever $R$ happens to emit, rather than by making
$R$ produce something readable. We supervise $R$ directly against the cross-attention's own
output under the teacher caption. Since the self-attention that precedes it is identical in
both cases, it cancels, and only one forward pass is required: at each block we record
$o_i^t = X(h_i, \tau)$ with the encoded caption $\tau$, then recompute the same layer with
the predicted tokens, $o_i^s = X(h_i, R(h_i, i, t))$, and minimise
\begin{equation}
    \mathcal{L} = \frac{1}{N}\sum_{i=1}^{N} \big\| o_i^s - o_i^t \big\|^2 .
\end{equation}
This gives a dense signal at every block and every timestep, far stronger than a gradient
routed through the full diffusion objective, and lets $R$ reach a useful regime quickly.

\subsection{Stage 3: Caption-Free Finetuning}
\label{sec:stage3}

Stage-2 trains $R$ to imitate the caption pathway, which caps it at what that pathway
provided. In Stage-3 we remove the caption entirely and let the model improve beyond it
under the generation objective alone. $R$ is initialised from Stage-2 and remains a
single module shared across all $N$ blocks, taking $i$ and $t$ as inputs as before, and
is finetuned at a lower learning rate than in Stage-2.
The self-attention layers are finetuned further, so the backbone can adapt to a
conditioning signal that is now derived from its own hidden states rather than from
text. The cross-attention layers stay frozen. This is deliberate and load-bearing: if
they could adapt, the loss could be reduced by moving the cross-attention toward whatever
$R$ happens to emit. Training uses the same
flow-matching objective as Stage-1, again supervised on the egocentric half of the canvas
only, with $\tau$ replaced by $R(h_i, i, t)$ at every block.

\subsection{Inference}
\label{sec:inference}

At inference the model requires only an exocentric video and a target head trajectory.
The trajectory is used to reproject the depth estimate into an egocentric prior. At every block, $R$ produces the conditioning tokens from the current hidden
state, so the description the cross-attention reads is rebuilt at each step as the
egocentric view emerges. No caption is written, no language model or text encoder is
loaded, and no geometry-guided attention bias is computed.

\section{Experiments}

\paragraph{Dataset.}
We train and evaluate on Ego-Exo4D~\citep{grauman2024ego}, which provides synchronised
egocentric and exocentric recordings with camera poses across a range of everyday
activities. We apply a motion-aware curation step to select clips with sufficient head motion, described in Appendix~\ref{app:curation}. Each clip is $25$ frames at $448\!\times\!448$ for the egocentric view and
$448\!\times\!768$ for the exocentric one. Unlike prior work, which restricts training to
the single best-reconstructed exocentric camera per take, We retain all available exocentric viewpoints, yielding {100K} clips for training and {400} clips for evaluation on seen and unseen scenes. Captions for Stage-1 are generated with
Qwen3-VL-32B-Instruct~\citep{Qwen3-VL}, prompted to describe the exocentric and
egocentric views in separate blocks (Appendix~\ref{app:prompt}); the egocentric prior is obtained by estimating depth
with MoGe-v2~\citep{wang2025moge2} and reprojecting along the ground-truth
head trajectory.

\paragraph{Implementation.}
Our backbone is Wan2.1-I2V-14B~\citep{wan2025}, with $N=40$ transformer blocks and an inner
dimension of $5120$. Across all stages we adapt only the attention projections, with LoRA
of rank $r=128$ and $\alpha=128$, together with the input patch embedding. The Dynamic
Captioner uses $n=256$ queries per half, giving $512$ conditioning tokens, which matches
the sequence length the backbone's cross-attention was pretrained on; it has
238M parameters in total. Stage-1 is trained for \textbf{1} epoch with AdamW~\citep{loshchilov2017decoupled} at a learning rate of
$5\!\times\!10^{-5}$. Stage-2 freezes the Stage-1 weights and trains only the Dynamic
Captioner for \textbf{0.66} epoch at $1\!\times\!10^{-5}$, supervising against the frozen
cross-attention's output at all blocks per step. Stage-3
finetunes the Dynamic Captioner at $5\!\times\!10^{-6}$ and the self-attention LoRA at
$1\!\times\!10^{-5}$ for \textbf{1} epoch, with the cross-attention frozen throughout.
All stages use a batch size of $1$ with gradient checkpointing on a single
{H200} GPU, a flow-matching schedule with $1000$ training timesteps, and a
classifier-free guidance dropout rate of $0.1$. During inference, we sample with the flow-matching Euler scheduler for $40$ steps at a guidance scale of
$3.0$. Training duration and computational resources for each phase are reported in Appendix~\ref{app:training}.

\paragraph{Metrics.}
Following~\citep{kang2026egox}, we report PSNR, SSIM, LPIPS and
CLIP-I between each generated frame and its ground truth. We additionally report FVD~\citep{ge2024content} and the three temporal
measures of VBench~\citep{zhang2025evaluation} --- Temporal Flickering, Motion Smoothness and
Dynamic Degree.

\subsection{Quantitative and Qualitative Results}
\begin{figure}[t]
    \includegraphics[width=1.0\textwidth]{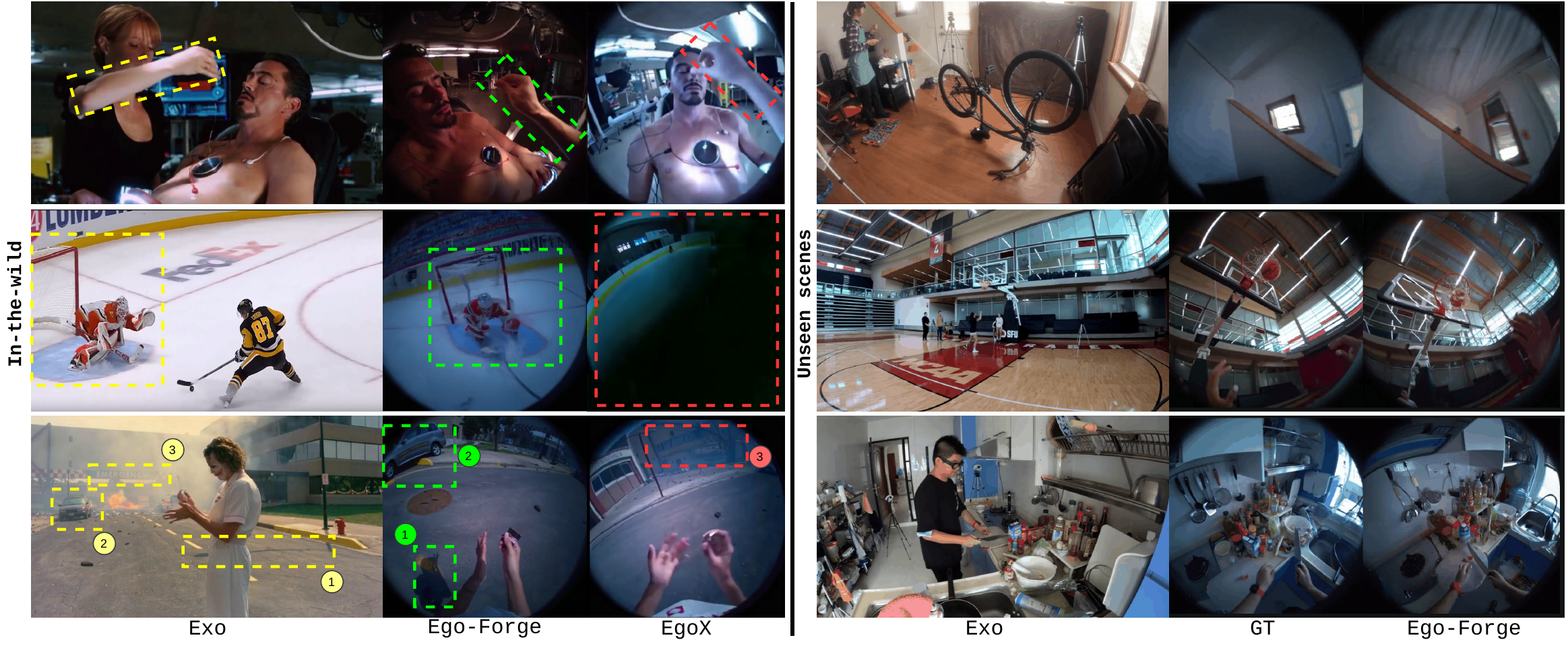}
    \caption{\textbf{Qualitative results.} Left: a) Ours better positions the arm on the man's chest. b) Ours handles small exo-camera motion. c) \underline{1}: Ours uses visual cues, e.g., the exo-view shadow. \underline{2}: Ours better synthesizes unseen content, using the car observed in the exo view. \underline{3}: The baseline misinterprets the ``EMERGENCY'' sign (mentioned in caption) and generates it at the wrong time. }

    \label{fig:main_vis}
\end{figure}

As shown in Table~\ref{tab:main}, our method achieves the best overall performance on both seen and unseen scenes~\citep{kang2026egox}. During data curation, we ensure that these scenes do not overlap with the training set. Figure~\ref{fig:main_vis} compares our method with EgoX on in-the-wild examples, while also showcasing our results on unseen scenes. More visualizations are provided in Appendix~\ref{app:ex_vis}. Figure~\ref{fig:speed} reports the training size and inference cost of both models, with inference performed on 25 frames.

\begin{table*}[t]
\centering
\small
\setlength{\tabcolsep}{1pt}
\begin{tabular}{llccccccccc}
\toprule
 & & \multicolumn{4}{c}{Image Criteria} & \multicolumn{4}{c}{Video Criteria} \\
\cmidrule(lr){3-6}\cmidrule(lr){7-10}
Scenes & Method & PSNR $\uparrow$ & SSIM $\uparrow$ & LPIPS $\downarrow$ & CLIP-I $\uparrow$
 & FVD $\downarrow$ & \makecell{Temporal\\Flickering} $\uparrow$
 & \makecell{Motion\\Smoothness} $\uparrow$ & \makecell{Dynamic\\Degree} $\uparrow$ \\
\midrule
\multirow{7}{*}{Seen}
 & Exo2Ego-V           & {14.41} & 0.382 & {0.564} & 0.794 & 642.09 & 0.953 & 0.944 & \underline{0.986} \\
 & Trj-Crafter & 13.59 & 0.397 & 0.591 & 0.788 & 741.72 & 0.960 & 0.980 & 0.950 \\
 & Wan-FCtrl      & 13.11 & {0.433} & 0.622 & 0.789 & 610.13 & 0.966 & 0.980 & 0.906 \\
 & Wan VACE             & 13.48 & 0.453 & 0.611 & {0.771} & 583.29 & \textbf{0.989} & \textbf{0.994} & 0.691 \\
 & Syn2Seq-F      & 15.01 & 0.461 & 0.549 & 0.793 & 513.73 & 0.965 & 0.970 & 0.822 \\
 & EgoX                   & \underline{16.03} & \underline{0.555} & \underline{0.498} & \textbf{0.901} & \underline{189.71} & \underline{0.975} & {0.991} & {0.969} \\
 & \textbf{Ego-Forge (Ours)}             & \textbf{16.37} & \textbf{0.557} & \textbf{0.492} & \underline{0.893} & \textbf{170.18} & \textbf{0.989} & \underline{0.992} & \textbf{0.990} \\
\midrule
\multirow{7}{*}{Unseen}
 & Exo2Ego-V           & 12.93 & {0.430} & {0.594} & 0.677 & 1123.90 & 0.966 & 0.970 & {0.978} \\
 & Trj-Crafter & 12.44 & 0.301 & 0.619 & 0.768 & {803.11} & 0.966 & 0.984 & 0.940 \\
 & Wan-FCtrl        & {13.13} & 0.439 & 0.615 & 0.790 & 960.28 & 0.971 & 0.985 & 0.944 \\
 & Wan VACE             & 12.97 & 0.345 & 0.638 & {0.820} & 1023.19 & \textbf{0.995} & \textbf{0.996} & 0.427 \\
 & Syn2Seq-F      & 13.63 & 0.402 & 0.617 & 0.715 & 891.43 & 0.933 & 0.944  & 0.733 \\
 & EgoX                     & \underline{14.32} & \underline{0.455} & \underline{0.555} & \textbf{0.891} & \underline{463.24} & \underline{0.984} & \underline{0.990} & \textbf{0.987} \\
 & \textbf{Ego-Forge (Ours)}             & \textbf{14.60} & \textbf{0.484} & \textbf{0.533} & \underline{0.864} & \textbf{313.95} & \underline{0.984} & 0.981 & \underline{0.979} \\
\bottomrule
\end{tabular}
\caption{\textbf{Quantitative comparison on Ego-Exo4D.} \textbf{Bold} and \underline{underlined} denote the best and second-best results, respectively. Trj-Crafter~\citep{yu2025trajectorycrafter}, Wan-FCtrl~\citep{videoxfun2024}, Wan VACE~\citep{jiang2025vace}, and Syn2Seq-F~\citep{mahdi2026synchrony}.}

\label{tab:main}
\end{table*}


\subsection{Ablation Study}

\begin{figure*}[t]
\centering

\begin{minipage}[c]{0.68\textwidth}
\centering
\small
\setlength{\tabcolsep}{2pt}

\begin{tabular}{lccccc}
\toprule
Conditioning & PSNR $\uparrow$ & SSIM $\uparrow$ & LPIPS $\downarrow$ & CLIP-I $\uparrow$ & FVD $\downarrow$ \\
\midrule
None (zeroed cross-attention)
& 14.87 & 0.469 & 0.556 & 0.772 & 360.46 \\
Text caption
& 16.24 & 0.542 & 0.498 & \textbf{0.901} & 187.73 \\
\textbf{Dynamic Captioning}
& \textbf{16.37} & \textbf{0.557} & \textbf{0.492} & 0.893 & \textbf{170.18} \\
\bottomrule
\end{tabular}

\captionof{table}{\textbf{The cross-attention input.}
Removing the caption is only worthwhile if something replaces it:
the first row leaves the cross-attention empty, which isolates the
contribution of our module.}
\label{tab:abl_conditioning}

\end{minipage}
\hfill
\begin{minipage}[c]{0.3\textwidth}
\centering
\includegraphics[width=0.8\textwidth]{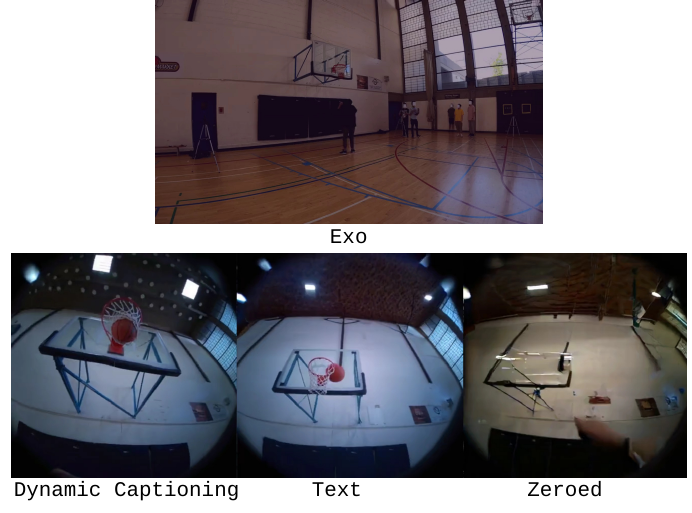}
\centering
\captionof{figure}{\textbf{Examples for the left. } Better viewed zoomed.}
\label{fig:placeholder}

\vspace{16pt}
\end{minipage}

\end{figure*}

We perform a series of ablation studies to validate the necessity of Dynamic Captioning and its components. As shown in Table~\ref{tab:abl_conditioning}, removing the textual stream substantially degrades performance (Figure~\ref{fig:placeholder}), while text captions provide a plausible teacher signal for Stage-2 training. Our method first learns from this teacher in Stage 2 and subsequently surpasses it through Stage 3.

\begin{figure*}[t]
\centering

\begin{minipage}[c]{0.68\textwidth}
\centering
\small
\setlength{\tabcolsep}{3pt}

\begin{tabular}{ccccccc}
\toprule
Block $i$ & Timestep $t$ & PSNR $\uparrow$ & SSIM $\uparrow$ & LPIPS $\downarrow$ & CLIP-I $\uparrow$ & FVD $\downarrow$ \\
\midrule
\xmark & \xmark & 13.93 & 0.391 & 0.601 & 0.690 & 540.78 \\
\xmark & \cmark & 14.67 & 0.444 & 0.576 & 0.739 & 375.75 \\
\cmark & \xmark & 14.88 & 0.490 & 0.549 & 0.775 & 281.42 \\
\cmark & \cmark & \textbf{16.37} & \textbf{0.557} & \textbf{0.492} & \textbf{0.893} & \textbf{170.18} \\
\bottomrule
\end{tabular}

\captionof{table}{\textbf{Conditioning the Dynamic Captioner.}
The block index lets one shared module specialise by depth; the timestep lets it shift toward the egocentric half as the view emerges (Figure~\ref{fig:ego_attention}).}
\label{tab:abl_conditioning_it}

\end{minipage}
\hfill
\begin{minipage}[c]{0.3\textwidth}
\centering
\includegraphics[width=0.8\textwidth]{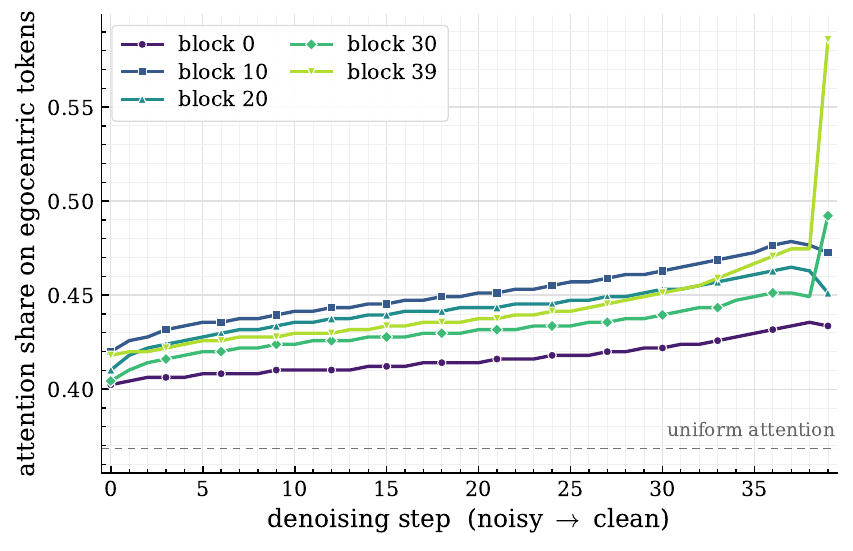}
\centering
\captionof{figure}{\textbf{Dynamic Captioner Attention.} The dashed line is the share that uniform attention would give.}
\label{fig:ego_attention}

\vspace{10pt}
\end{minipage}

\end{figure*}

Table~\ref{tab:abl_conditioning_it} compares Dynamic Captioning with and without the block index and timestep as conditioning signals. We find that removing the block index $i$ degrades performance the most, as the model then lacks an explicit notion of depth across DiT blocks, causing most blocks to receive out-of-distribution inputs. Figure~\ref{fig:ego_attention} shows where the Dynamic Captioner looks as denoising proceeds. Every block shifts its attention toward the egocentric half as the view emerges, and the effect grows sharply with depth: the last block shifts by $0.17$ against the first block's $0.03$. This is what the timestep and block-index conditioning were designed to allow, and it is not available to a fixed caption, which is read identically at every block and every step. 

\begin{figure*}[h]
\centering

\begin{minipage}[c]{0.68\textwidth}
\centering
\small
\setlength{\tabcolsep}{2pt}

\begin{tabular}{llcccc}
\toprule
Module & \makecell{Visual\\features} & PSNR $\uparrow$ & SSIM $\uparrow$ & LPIPS $\downarrow$ & CLIP-I $\uparrow$ \\
\midrule
Distribution matching & CLIP + DINO & 14.06 & 0.388 & 0.611 & 0.702 \\
\textbf{Dynamic Captioning} & VAE latents & \textbf{16.37} & \textbf{0.557} & \textbf{0.492} & \textbf{0.893} \\
\bottomrule
\end{tabular}

\captionof{table}{\textbf{Comparison to a static captioner.} The static captioner projects CLIP~\citep{radford2021clip} and DINO~\citep{oquab2024dinov2} features into the text-token space using distribution-matching losses. (Appendix~\ref{app:adapter})}

\label{tab:abl_adapter}

\end{minipage}
\hfill
\begin{minipage}[c]{0.3\textwidth}
\centering
\includegraphics[width=1.0\textwidth]{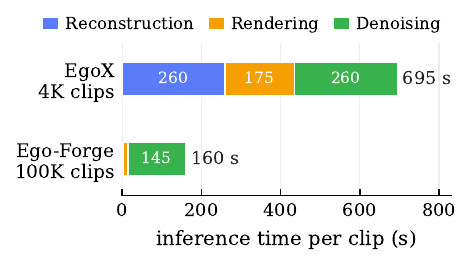}
\centering
\captionof{figure}{\textbf{Training size and inference cost comparison.}}
\label{fig:speed}

\vspace{10pt}
\end{minipage}

\end{figure*}

Furthermore, Table~\ref{tab:abl_adapter} compares Dynamic Captioning with a Static Captioning module, where an adapter maps exocentric visual tokens to the textual-token distribution and produces a static embedding for the cross-attention layers (further details are provided in Appendix~\ref{app:adapter}). Finally, Table~\ref{tab:abl_training} shows that Stage 3 further improves upon Stage 2, yielding both better quantitative results and higher-quality generations (Figure~\ref{fig:training}).




\begin{figure*}[!h]
\centering

\begin{minipage}[c]{0.68\textwidth}
\centering
\small
\setlength{\tabcolsep}{1pt}

\begin{tabular}{lccccc}
\toprule
Training & PSNR $\uparrow$ & SSIM $\uparrow$ & LPIPS $\downarrow$ & CLIP-I $\uparrow$ & FVD $\downarrow$ \\
\midrule
Stage 2 only (distillation)
& 15.27 & 0.498 & 0.510 & 0.827 & 243.46 \\
\textbf{+ Stage 3 (joint finetuning)}
& \textbf{16.37} & \textbf{0.557} & \textbf{0.492} & \textbf{0.893} & \textbf{170.18} \\
\bottomrule
\end{tabular}

\captionof{table}{\textbf{Distillation alone against joint finetuning.}
Stage-3 removes the caption entirely and let the model improve beyond its teacher.}
\label{tab:abl_training}

\end{minipage}
\hfill
\begin{minipage}[c]{0.3\textwidth}
\centering
\includegraphics[width=1.0\textwidth]{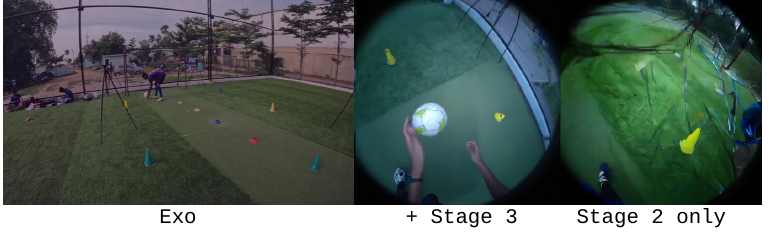}
\centering
\captionof{figure}{\textbf{ Examples for the left.} Better viewed zoomed.}
\label{fig:training}

\vspace{16pt}
\end{minipage}

\end{figure*}


\section{Conclusion}
We presented \textit{\textbf{Ego-Forge}}, an exo-to-ego video generation model that eliminates the textual and geometric conditioning required by prior work. To avoid the need for a caption describing the non-existent egocentric view, we introduce \textit{Dynamic Captioning}: a single module shared across network blocks and conditioned on network depth and diffusion timestep, which produces conditioning tokens directly from the model's hidden states. Through a staged training procedure, the module first learns to reproduce the conditioning provided by the pretrained text pathway and is then jointly refined with the diffusion model without text. We further show that geometric attention can be removed by scaling training to a large dataset spanning all available exocentric viewpoints, allowing the model to learn cross-view correspondence directly from data while avoiding the propagation of reconstruction errors. Together, these design choices yield a simpler and more efficient pipeline that requires only the exocentric video and a lightweight geometric prior at inference. Ego-Forge outperforms the state of the art on the Ego-Exo4D benchmark while running $4.3\times$ faster end-to-end, requires no human-written caption at inference, and generalises to in-the-wild footage, including challenging cases where geometric correspondence alone is insufficient to recover appearance.

\subsection*{AI use statement}

We used generative AI in this work for generating synthetic data, drafting parts
of the paper, polishing writing, and assisting with code. The textual captions
used to condition Stage~1 training were produced automatically with
Qwen3-VL-32B-Instruct. Initial drafts of parts of the method were produced with a large language model and then substantially
revised by the authors. Parts of the training, evaluation and visualisation code were written with AI assistance, and all of it was read, executed and verified by the authors.

\bibliography{iclr2027_conference}
\bibliographystyle{iclr2027_conference}

\clearpage
\setcounter{figure}{0}\renewcommand{\thefigure}{\Alph{figure}}
\setcounter{table}{0}\renewcommand{\thetable}{\Alph{table}}

\phantomsection
\addcontentsline{toc}{section}{Appendix}
\begin{center}
  {\Large\sc Appendix}
\end{center}

\appendix

\section{Textual and visual suppression by GGA}
\label{app_teaser}
\begin{figure}[h]
    \centering
    \includegraphics[width=\linewidth]{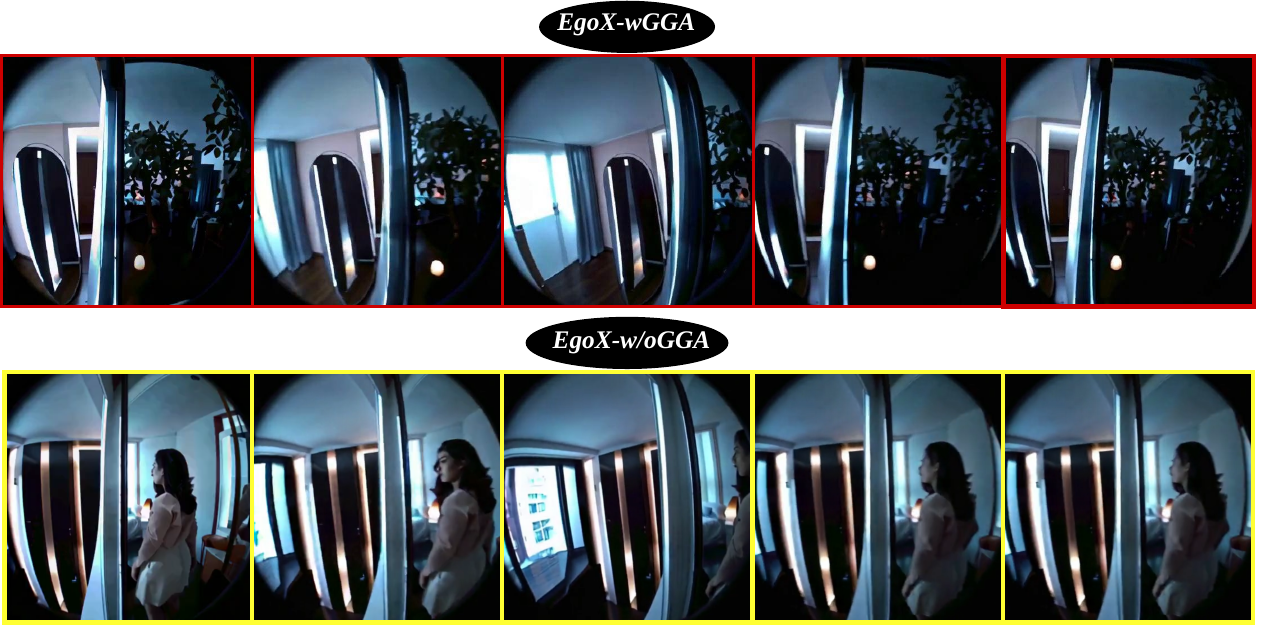}
    \caption{\textbf{EgoX~\citep{kang2026egox} performance depending on GGA presence.}}
    \label{app:ts}
\end{figure}

As illustrated in Figure~\ref{app:ts}, removing GGA allows EgoX to recover the person's reflection in the generated video, although the reflection remains inaccurate and the model still fails to produce the correct egocentric view.

\section{Training data curation}
\label{app:curation}

Ego-Exo4D provides long recordings, but not every window within them is useful for
our task. A clip in which the head barely moves gives the model little to learn
from: the egocentric prior is nearly static, the target view changes little, and
the correspondence between the two views is trivial. We therefore select windows by
the amount of head motion they contain.

\paragraph{Motion metric.}
For a window of $F$ frames with egocentric extrinsics $\{[R_i \mid t_i]\}_{i=1}^{F}$,
we measure the motion between consecutive frames as the sum of a translational and
a rotational term,
\begin{equation}
    m_i = \lVert t_{i+1} - t_i \rVert_2
        + \lambda_{\text{rot}} \cdot
          \arccos\!\left(\frac{\operatorname{tr}(R_i^\top R_{i+1}) - 1}{2}\right),
\end{equation}
where the second term is the geodesic angle of the relative rotation. We set
$\lambda_{\text{rot}} = 1$, so that one radian of rotation is weighted equally with
one metre of translation; Ego-Exo4D poses are metric, and over a short window the
two terms are of comparable magnitude.

\paragraph{Selection.}
We summarise each window by the mean of $m_i$ and retain those above a threshold,
discarding windows in which the head is effectively stationary. We additionally
compute a robust within-window statistic: taking $\psi = \operatorname{med}(m) +
2\operatorname{MAD}(m)$, the fraction of frames exceeding $\psi$ measures how
\emph{bursty} the motion is rather than how large it is, and lets us identify
windows whose motion is concentrated in a few frames.

\paragraph{Category-balanced ordering.}
Ego-Exo4D groups takes by activity, and the categories are very unevenly sized.
Read in order, a training run would see long runs of a single activity, which at
batch size~1 risks the model drifting toward whichever category it is currently
inside. We therefore group clips by activity, shuffle within each group, and then
draw round-robin across groups, so that consecutive samples come from different
activities. Groups are exhausted at different rates and drop out as they empty; no
clip is repeated or discarded, so the ordering changes only the sequence in which
the data is seen. Unlike prior work, we retain all exocentric viewpoints per take
rather than only the best-reconstructed one, which is what brings the training set
to 100k clips.

\section{Caption generation pipeline}
\label{app:prompt}

Stage~1 conditions on a caption describing both views. Ego-Exo4D provides no such
annotation, so we generate one for every training clip with
Qwen3-VL-32B-Instruct~\citep{Qwen3-VL}, an open-weight vision--language model. Using
an open model rather than a paid API is what makes captioning a set of this size
feasible; it is also the cost our method removes entirely, since Ego-Forge needs no
caption at inference.

\paragraph{Inputs and ordering.}
The model receives both videos of a clip in a single call, with the \emph{egocentric}
view first. The ordering matters: the two views are passed as one token sequence, so
if the budget is exceeded the tail is truncated, and we prefer to lose exocentric
detail rather than egocentric. For the same reason the prompt asks for roughly
$150$ words of egocentric description against $50$ of exocentric --- the egocentric
block is what the model is being asked to generate, and the exocentric view is
already supplied to it as pixels.

\paragraph{Prompt design.}
The prompt is given verbatim in Listing~\ref{lst:prompt}. Four of its directives
address failure modes we observed in earlier iterations:

\begin{itemize}
\item \textbf{View separation.} Without an explicit instruction, the model freely
described objects visible only in the exocentric view inside the egocentric block.
Since that block is meant to describe what the wearer sees, such content is
information the exocentric camera has but the wearer does not, and it makes the
caption inconsistent with the target. The prompt forbids it, requires first person
in the egocentric block, and forbids describing the wearer's own body.

\item \textbf{Specificity.} Early captions leaned on generic nouns --- ``tool'',
``device'', ``utensil'' --- which carry almost no conditioning signal. The prompt
requires precise names together with colour, material and position.

\item \textbf{Objectivity.} Subjective adjectives (``modern'', ``cluttered'') describe
the annotator rather than the scene. The prompt restricts descriptions to physical,
observable attributes.

\item \textbf{Honest abstention.} Asked to describe every clip in detail, the model
invented activity for clips in which little happens or the hands are out of frame.
The prompt instructs it to say so plainly and keep the block short instead.
\end{itemize}

\paragraph{Filtering.}
Generation is sharded across GPUs and cached per clip. We then audit the output
automatically for two failure modes: captions that are empty or truncated
mid-sentence, and captions whose two blocks are near-duplicates of one another,
which indicates the view separation failed. Clips failing either check are excluded
from Stage~1 training.

\paragraph{Output format.}
According to EgoX~\citep{kang2026egox}, captions follow a fixed two-block structure, each block carrying a static scene
overview followed by an action analysis:

\begin{quote}\small
\texttt{[Ego view] Scene Overview: \dots{} Action Analysis: \dots{}}\\
\texttt{[Exo view] Scene Overview: \dots{} Action Analysis: \dots{}}
\end{quote}

\section{Training budget}
\label{app:training}

All training was done on NVIDIA~H200 GPUs. Table~\ref{tab:budget} gives the cost of
each stage.

\begin{table}[h]
\centering
\small
\begin{tabular}{lccl}
\toprule
Stage & GPUs & Days & Trainable \\
\midrule
1: Text-conditioned adaptation & 2 & 5 & LoRA on $S$, $X$; patch embedding \\
2: Learning the Dynamic Captioner & 1 & 3 & $R$ only \\
3: Caption-free finetuning & 4 & 2 & $R$ and LoRA on $S$ \\
\bottomrule
\end{tabular}
\caption{\textbf{Training budget.} All stages run at batch size~1 per GPU with
gradient checkpointing; the $14$B backbone and the stitched canvas leave little
memory for a larger batch.}
\label{tab:budget}
\end{table}

\paragraph{Preprocessing.}
Beyond training, the dataset requires two one-time passes: depth estimation and
reprojection to produce the egocentric priors, and caption generation with
Qwen3-VL-32B-Instruct for Stage~1. Both are sharded across GPUs and cached.

\section{Static Captioning}
\label{app:adapter}

\paragraph{Overview.} We term our exocentric feature adapter the \emph{Static Captioner}. Its role is to translate exocentric visual tokens into the text-embedding space that the diffusion backbone's cross-attention was pretrained to consume, so that visual conditioning can be injected through the same pathway originally trained on text, without any accompanying caption at inference time.

\paragraph{Architecture and inputs.} Each exocentric clip is encoded with frozen CLIP and DINOv2 vision encoders. Their patch features are concatenated to form a sequence of tokens of dimension $2048$ (CLIP $1024$ $\oplus$ DINOv2 $1024$). The Static Captioner is a lightweight trainable module that maps these visual tokens into $4096$ dimension in the umT5 text-embedding space. These projected visual tokens are the \emph{only} input to the backbone's cross-attention layers; no textual caption is used at inference.

\paragraph{Pretraining data.} We pretrain the Static Captioner on an external, diverse corpus rather than our egocentric training set, so that the learned visual-to-text mapping generalizes beyond a single domain. Starting from OpenVid-1M~\citep{nan2025openvid}, we apply caption-length, resolution, aspect-ratio, and frame-count filters, yielding approximately $200\text{K}$ video–caption pairs. All frames are motion-normalized to a common frame rate before feature extraction, and CLIP/DINOv2/umT5 features are precomputed and cached.

\paragraph{Distribution-alignment objective.} A caption is a linguistically \emph{ordered} sequence, whereas our visual tokens are \emph{spatially} ordered; enforcing a per-token correspondence would therefore destroy the spatial layout our conditioning relies on. We instead align the two token sets at the \emph{distribution} level, which is order-agnostic and thus preserves the spatial identity and ordering of the visual tokens. For a clip, let $X=\{x_i\}_{i=1}^{N}$ be the adapter's projected visual tokens and $Y=\{y_j\}_{j=1}^{M}$ the valid umT5 caption tokens, both in $\mathbb{R}^{d}$. We consider three alignment objectives.

\emph{Mean alignment} matches only the first-order statistics of the two distributions, i.e. their centroids:
\begin{equation}
\mathcal{L}_{\text{mean}}(X,Y) = 1 - \cos\big(\bar{x},, \bar{y}\big),
\qquad \bar{x} = \frac{1}{N}\sum_i x_i,\quad \bar{y} = \frac{1}{M}\sum_j y_j .
\end{equation}

\emph{CORAL} additionally matches second-order statistics by aligning the feature covariances of the two token clouds:
\begin{equation}
\mathcal{L}_{\text{coral}}(X,Y) = \frac{1}{d^2},\big| C_X - C_Y \big|_F^2 ,
\end{equation}
where $C_X$ and $C_Y$ are the covariance matrices of $X$ and $Y$ and $\|\cdot\|_F$ is the Frobenius norm.

\emph{Sliced Wasserstein Distance (SWD)} approximates the optimal-transport distance between the two distributions by projecting the tokens onto random one-dimensional directions and matching their sorted marginals:
\begin{equation}
\mathcal{L}{\text{swd}}(X,Y) = \frac{1}{L}\sum{\ell=1}^{L} W_2^2!\Big( {\theta_\ell^{\top} x_i}i,; {\theta\ell^{\top} y_j}_j \Big),
\end{equation}
where each $\theta_\ell$ is a random unit direction and $W_2$ is the one-dimensional Wasserstein-2 distance (computed by sorting).

Minimizing any of these objectives drives the \emph{distribution} of visual tokens toward that of the caption tokens, so that individual visual tokens become compatible with the text-trained cross-attention while retaining their spatial layout. Because all three are computed over token \emph{sets} rather than aligned pairs, they impose no linguistic ordering on the visual tokens.

\section{Additional Visualizations}
\label{app:ex_vis}
Additional qualitative results are provided in the supplementary materials ZIP file. The accompanying videos contain more extensive visualizations of our method, where the temporal evolution and flow of the generated egocentric videos can be better appreciated than from individual frames.

\begin{lstlisting}[style=promptstyle, caption={The prompt used to caption every
training clip with Qwen3-VL-32B-Instruct. Both videos are passed in a single call,
egocentric first.}, label={lst:prompt}, float=*t]
You are a hyper-realistic scene reconstruction AI. You are given TWO videos of the same moment. The FIRST video is the first-person (egocentric) view, recorded by a camera worn on the head of the person performing the activity. The SECOND video is the third-person (exocentric) view of that same person. Analyze both and produce a two-part analysis for each: a static scene overview followed by a dynamic action breakdown. Your guiding principle is strict objectivity.

--- MISSION PROTOCOL ---
Phase 1: Scene Establishment
First, analyze all provided frames to establish a detailed, static description of the physical environment. Detail the surfaces (walls, floors), furniture, and all unmoving background items. This is your 'establishing shot'.
Phase 2: Action Transition Analysis
After establishing the scene, provide a detailed description of the action progression and transitions observed across the sequence. Focus on how actions evolve, change, and flow from one moment to the next, maintaining awareness of the overall context established in Phase 1.

--- CRITICAL DIRECTIVES ---
1. Exhaustive Object Inventory: THIS IS YOUR MOST IMPORTANT TASK. You must meticulously identify and catalog EVERY visible item.
- NO GENERIC TERMS: Do not use vague words like 'tool', 'box', 'utensil', or 'device'.
- BE SPECIFIC: Use precise names (e.g., 'smartphone', 'coffee mug', 'wooden spoon', 'cutting board', 'refrigerator', 'laptop computer', 'ceramic bowl', 'stainless steel knife').
- DESCRIBE PROPERTIES: Include colors, materials, textures, and positions (e.g., 'a blue ceramic mug on a granite countertop').
2. Focus on Hand-Object Interaction: THE ACTION'S CORE.
- Your primary narrative focus MUST be the hands. Describe their precise posture, movement, and interaction with objects (e.g., 'the right hand grasps the knife handle,' 'the left hand's fingertips stabilize the tomato').
- Every action description should revolve around what the hands are doing.
3. Strict Objectivity: DESCRIBE, DO NOT INTERPRET.
- AVOID JUDGMENT: Do not use subjective or abstract adjectives (e.g., AVOID 'modern', 'beautiful', 'cluttered', 'well-lit'). Describe only physical, measurable attributes.
4. Transition-Focused Analysis
- Analyze the sequence as a continuous flow of actions
- Describe how movements and interactions transition and evolve
- Focus on the progression and changes rather than individual frame descriptions
- Maintain narrative continuity throughout the sequence
5. VIEW SEPARATION: THE TWO BLOCKS MUST NOT REPEAT EACH OTHER.
- The [Ego view] block must be written ONLY from the FIRST video. Describe only what falls inside that camera's frame: the hands, the objects they touch or look at, and the surface directly in front of them.
- If something appears only in the SECOND video, it must NOT appear anywhere in the [Ego view] block.
- In the [Ego view] block, write in the first person. Never write 'the person', and never describe the wearer's own body, face, or clothing.
- Never mention cameras, tripods, or recording equipment in either block.
- If the video, especially the first-person view video, shows little activity or the hands are not visible, say so plainly and keep the block short. Do not invent action.

--- OUTPUT STRUCTURE ---
You MUST follow this exact two-block format:
[Ego view] Scene Overview: Detailed description of the static environment as seen in the FIRST video, from my own first-person perspective. List the objects within my field of view. Action Analysis: Describe the progression of actions and transitions throughout the sequence from my first-person perspective. Focus on how my hands move, how the objects I hold change, and the flow of the activity from beginning to end.
[Exo view] Scene Overview: Detailed description of the static background environment as seen in the SECOND video, from the third-person perspective. List all background objects. Action Analysis: Describe the progression of actions and transitions observed throughout the sequence. Focus on how movements evolve, interactions change, and the flow of activities from beginning to end.

--- LENGTH ---
Write about 150 words for the [Ego view] block and about 50 words for the [Exo view] block: 200 words in total. The [Ego view] block carries most of the detail. Both blocks must be complete.
\end{lstlisting}
\end{document}